# Thermal and RGB Images Work Better Together in Wind Turbine Damage Detection

**SERHII SVYSTUN[1], OLEKSANDR MELNYCHENKO[1], PAVLO RADIUK[2], OLEG SAVENKO[1], ANATOLIY SACHENKO[3,4], AND ANDRII LYSYI[1]**

[1]Department of Computer Engineering and Information Systems, Khmelnytskyi National University, 29016 Khmelnytskyi, Ukraine
[2]Department of Computer Science, Khmelnytskyi National University, 29016 Khmelnytskyi, Ukraine
[3]Research Institute for Intelligent Computer Systems, West Ukrainian National University, 46020 Ternopil, Ukraine
[4]Department of Informatics, Kazimierz Pulaski University of Technology and Humanities in Radom, 26600 Radom, Poland

Corresponding author: Serhii Svystun (e-mail: svystuns@khmnu.edu.ua).

**ABSTRACT** The inspection of wind turbine blades (WTBs) is crucial for ensuring their structural integrity and operational efficiency. Traditional inspection methods can be dangerous and inefficient, prompting the use of unmanned aerial vehicles (UAVs) that access hard-to-reach areas and capture high-resolution imagery. In this study, we address the challenge of enhancing defect detection on WTBs by integrating thermal and RGB images obtained from UAVs. We propose a multispectral image composition method that combines thermal and RGB imagery through spatial coordinate transformation, key point detection, binary descriptor creation, and weighted image overlay. Using a benchmark dataset of WTB images annotated for defects, we evaluated several state-of-the-art object detection models. Our results show that composite images significantly improve defect detection efficiency. Specifically, the YOLOv8 model's accuracy increased from 91% to 95%, precision from 89% to 94%, recall from 85% to 92%, and F1-score from 87% to 93%. The number of false positives decreased from 6 to 3, and missed defects reduced from 5 to 2. These findings demonstrate that integrating thermal and RGB imagery enhances defect detection on WTBs, contributing to improved maintenance and reliability.

## I. INTRODUCTION

The rapid expansion of wind energy as a sustainable power source has led to the widespread installation of wind turbines across diverse and often remote locations. Ensuring the efficient and uninterrupted operation of these wind turbines is critical, necessitating regular inspections to detect possible technical defects and mechanical damages [1, 2]. Traditional inspection methods, such as manual visual inspections and rope-access techniques, are time-consuming, risky, and often inadequate for thoroughly examining the complex structures of wind WTBs.

UAVs have emerged as a transformative technology in the field of wind turbine inspection, offering flexible systems that can adapt to changing environmental conditions and expand monitoring capabilities. UAVs are now widely used in various sectors, including tracking forest fires [3], monitoring crop yields [4], and inspecting green energy facilities [5], such as wind turbines [6, 7]. The use of UAVs significantly increases the efficiency of the monitoring process [8], providing high-resolution images and access to hard-to-reach areas [9], which are often inaccessible or hazardous for human inspectors.

Despite these advancements, wind turbine monitoring faces several challenges. Wind turbines are subjected to harsh environmental conditions, leading to various forms of degradation such as erosion, cracks, delamination, and lightning strikes. Timely detection of these defects is essential to prevent critical wind turbine failures or costly downtime. However, the sheer size and height of modern turbines, along with their complex blade geometries, make comprehensive inspections difficult [10].

Modern UAV technologies equipped with advanced imaging sensors have the potential to address these challenges. The integration of multispectral cameras, including thermal imaging and RGB cameras [11], provides additional information about the condition of the turbines. Thermal images can reveal temperature anomalies [12] that may indicate subsurface defects or material degradation, while RGB images offer high-resolution visual details of surface conditions. However, thermal cameras often have lower resolution and are sensitive to environmental conditions, leading to ambiguous results. Conversely, RGB cameras cannot detect temperature anomalies, which are critical for identifying hidden defects not visible on the surface [13].

The limitations of using thermal or RGB cameras alone highlight the need for more sophisticated imaging solutions. Composite images that combine data from both types of cameras present a promising approach [14]. By fusing thermal and RGB imagery [15, 16], composite images provide detailed visualization due to the high resolution of RGB images and enable the detection of temperature anomalies inherent in thermal images.

At the same time, the development of deep learning (DL) methods has completely reshaped image processing and analysis [17, 18], offering powerful tools for automatic defect detection and classification. DL models, in particular convolutional neural networks (CNNs) [19–22], have demonstrated excellent performance in UAV image processing for a variety of applications, including object detection and semantic segmentation.

However, integrating thermal and RGB images poses challenges in terms of data alignment, fusion methodologies, and computational complexity [23]. Existing solutions often lack the ability to effectively combine multispectral data in a manner that fully leverages the strengths of each modality [24]. There is a need for novel methods that can seamlessly integrate thermal and RGB imagery and utilize DL frameworks to enhance defect detection accuracy.

In this study, we propose a multispectral image composition method that effectively combines thermal and RGB images captured by UAVs for enhanced defect detection on WTBs. By transforming spatial coordinates and applying advanced image fusion techniques, our method addresses the limitations of individual imaging modalities. We also leverage state-of-the-art DL models to process the composite images, improving detection accuracy and reliability.

The rest of the article is structured as follows. Section II reviews current research on wind turbine monitoring, UAV-based inspection systems, and the application of DL in UAV image processing. Section III formalizes the proposed method for composing multispectral images. Section IV presents the results of experimental tests with the proposed method, demonstrating its scientific value and practical significance through comparison with existing approaches. Section V briefly summarizes the main results obtained and outlines future research directions.

## II. RELATED WORKS

Recently, multispectral imaging systems have been actively employed in various critical industries, including infrastructure monitoring [25] and energy management [26]. The integration of UAVs significantly enhances the capabilities of these systems [27, 28], owing to UAVs' mobility and ability to access hard-to-reach areas [29]. This combination is especially beneficial for inspecting large structures like WTBs.

For instance, Zhang et al. [30] presented a method that combines high-resolution RGB imagery with thermal imaging to enhance defect detection in wind turbines. Although this approach capitalizes on the strengths of both imaging types, it mainly emphasizes data fusion, overlooking the challenges related to precise spatial alignment and integration essential for DL applications in wind turbine inspections.

Another work, Morando et al. [31], introduced a UAV-based inspection strategy for photovoltaic (PV) systems, employing both RGB and thermal cameras to track PV modules without GPS. While their technique improves inspection efficiency by optimizing flight paths and minimizing tracking errors, it is tailored specifically for PV plants and does not address the complexities of WTB inspections, such as blade curvature and diverse environmental conditions.

Zhou et al. [32] proposed an adaptive feature fusion module for RGB and infrared images, achieving significant detection accuracy improvements of up to 99%. However, their work is focused on general visual data processing, lacking a specific emphasis on wind turbine monitoring. The absence of considerations for UAV-acquired imagery and the distinctive challenges of turbine inspections limits the method's applicability in this domain.

Zhu et al. [33] addressed defect detection in WTBs through a multifunctional residual feature fusion network utilizing transfer learning. Although their method demonstrates potential, it relies extensively on pre-trained models and does not explore the fusion of UAV-captured thermal and RGB images. This gap may reduce its effectiveness in identifying a broader spectrum of defects, particularly those not easily identifiable through a single imaging modality.

Kwon et al. [34] proposed a calibration technique for active optical thermography, facilitating the detection of various WTB defects. While this approach advances thermal imaging, it lacks the integration of RGB data or sophisticated image processing methods, which could enhance assessment when used together. Similarly, Sanati et al. [35] examined passive and active thermography techniques to improve thermal image quality; however, their approach's accuracy and reliability are limited without incorporating RGB imagery and DL algorithms.

In their recent work, Memari et al. [36] successfully integrated multispectral imaging with ensemble learning for enhanced anomaly detection in small WTBs, improving detection accuracy through data fusion. Nevertheless, their study focuses on small turbines and does not address the challenges of scaling the approach to larger turbines or adapting to different environmental conditions, which are critical factors in practical, real-world applications.

In summary, taking into account the current limitations in the industry, this study aims to enhance defect detection in WTBs by developing a novel method that effectively composes multispectral images obtained from UAVs. By transforming spatial coordinates into a software coordinate model, optimizing errors, and minimizing them, our proposed method aims to enhance the accuracy and efficiency of inspections. The key scientific contributions of this study include:

- Developing an advanced image composition technique that seamlessly integrates thermal and RGB data from UAVs.
- Leveraging DL models to process the fused imagery for improved defect detection and classification.
- Demonstrating the method's effectiveness in real-world wind turbine inspections, addressing scalability and environmental challenges.

## III. METHOD

Thermal images provide critical information about temperature anomalies in green energy facilities. However, their resolution is lower compared to images obtained by RGB cameras. Combining data from both types aims to synchronise a new detailed state of the object model under study. The method's formalisation is presented in the scheme in Fig. 1.

### A. TRANSFORMATION OF SPATIAL COORDINATES

Block 1 of the proposed method involves transforming spatial coordinates into a software coordinate model five steps.

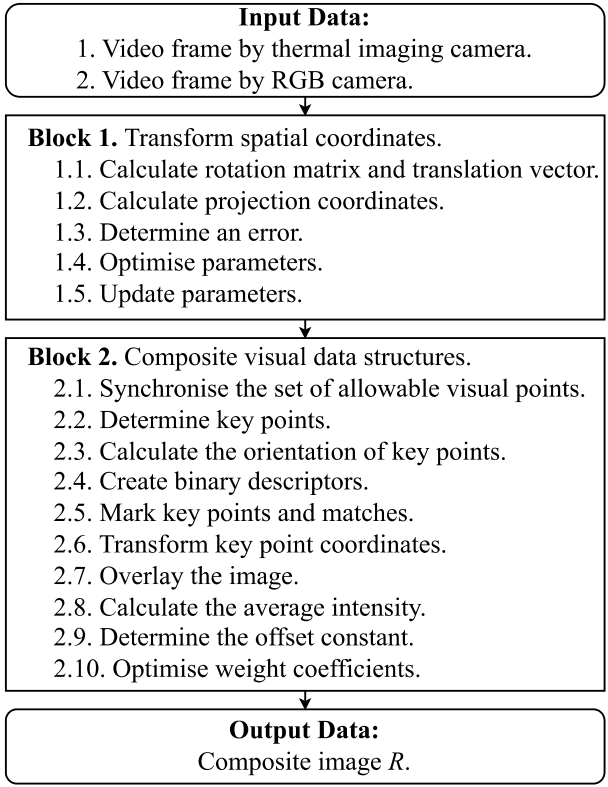

**Input Data:**
1. Video frame by thermal imaging camera.
2. Video frame by RGB camera.

**Block 1.** Transform spatial coordinates.
1.1. Calculate rotation matrix and translation vector.
1.2. Calculate projection coordinates.
1.3. Determine an error.
1.4. Optimise parameters.
1.5. Update parameters.

**Block 2.** Composite visual data structures.
2.1. Synchronise the set of allowable visual points.
2.2. Determine key points.
2.3. Calculate the orientation of key points.
2.4. Create binary descriptors.
2.5. Mark key points and matches.
2.6. Transform key point coordinates.
2.7. Overlay the image.
2.8. Calculate the average intensity.
2.9. Determine the offset constant.
2.10. Optimise weight coefficients.

**Output Data:**
Composite image $R$.

Figure 1. Scheme of the proposed multispectral image composition method that processes UAV video streams, transforms spatial coordinates, and composites visual data structures, ultimately producing a composite image $R$.

Step 1.1. Calculate the rotation matrix and translation vector. After receiving the input data, the rotation matrix $R$ and translation vector $t$ are calculated to integrate the spatial coordination of images from multiple sensors into a single coordinate system. It is aimed to minimise errors between the visual point coordinates of each UAV optical sensor.

Step 1.2. Calculate projection coordinates. The process of calculating projection coordinates $(u, v)$ is conducted using the current parameters. The spatial coordinates on the image plane are determined by the formulas:

$$u = f\frac{X_c}{Z_c} + c_u \quad \text{and} \quad v = f\frac{Y_c}{Z_c} + c_v, \qquad (1)$$

where $f$ is the focal length, and $(c_u, c_v)$ are the principal point coordinates, $X_c$ are the coordinates of the point in the camera coordinate system, and $X$ are the coordinates of the point in three-dimensional space.

Step 1.3. Determine an error. We determine the error $e_i$ as the difference between the actual coordinates of points on the image and the projection coordinates calculated by formula (1):

$$e_i = \sqrt{\left(u_i - u_i\right)^2 + \left(v_i - v_i\right)^2}, \qquad (2)$$

where $(u_i, v_i)$ are the actual coordinates of the point on the image and $(u_i, v_i)$ are the calculated coordinates of the point.

Step 1.4. Optimise parameters. The algorithmic sequence for calculating projection coordinates and determining errors requires optimising the parameters of the mathematical model to minimise error values. For this, an optimisation method using the Jacobian matrix [37], which contains partial derivatives of the errors concerning each model parameter, is applied. The formalisation of this process is given as follows:

$$J = \begin{pmatrix} \dfrac{\partial e_1}{\partial K_1} & \dfrac{\partial e_1}{\partial K_2} & \cdots & \dfrac{\partial e_1}{\partial D_1} & \cdots & \dfrac{\partial e_1}{\partial D_n} \\ \dfrac{\partial e_2}{\partial K_1} & \dfrac{\partial e_2}{\partial K_2} & \cdots & \dfrac{\partial e_2}{\partial D_1} & \cdots & \dfrac{\partial e_2}{\partial D_n} \\ \vdots & \vdots & \ddots & \vdots & \ddots & \vdots \\ \dfrac{\partial e_N}{\partial K_1} & \dfrac{\partial e_N}{\partial K_2} & \cdots & \dfrac{\partial e_N}{\partial D_1} & \cdots & \dfrac{\partial e_N}{\partial D_n} \end{pmatrix}, \qquad (3)$$

where $\left(\dfrac{\partial e_i}{\partial K_j}\right)$ and $\left(\dfrac{\partial e_i}{\partial D_k}\right)$ are the partial derivatives of error $e_i$ with respect to the model parameters $K_j$ and $D_k$.

The error $e_i$ from formula (2) for each point is determined as the difference between the three-dimensional coordinates and the coordinates obtained using the current model. The partial derivative of error $e_i$ to parameter $K_j$ shows how much the value of $e_i$ changes with a small change in parameter $K_j$. At the same time, other parameters remain unchanged and are defined as follows:

$$\frac{\partial e_i}{\partial K_j} = \lim_{\Delta K_j \to 0} \left( \frac{e_i\left(K_1, K_2, \ldots, K_j + \Delta K_j, \ldots, K_m\right)}{\Delta K_j} - \right.$$
$$\left. - \frac{e_i\left(K_1, K_2, \ldots, K_j, \ldots, K_m\right)}{\Delta K_j} \right). \qquad (4)$$

Similarly, the partial derivative for $D_k$ is determined.

The numerical calculation of the partial derivative concerning the elements of the rotation matrix $R$ is performed, according to the formula:

$$\frac{\partial e_i}{\partial R_{jk}} \approx \frac{e_i\left(R_{jk} + \Delta R_{jk}\right) - e_i\left(R_{jk}\right)}{\Delta R_{jk}},$$

where $\Delta R_{jk}$ is a small change in element $R_{jk}$.

The numerical representation of the partial derivative concerning the elements of the translation vector t is carried out as follows:

$$\frac{\partial e_i}{\partial t_k} \approx \frac{e_i\left(t_k + \Delta t_k\right) - e_i\left(t_k\right)}{\Delta t_k},$$

where $\Delta t_k$ is a small change in element $t_k$.

Step 1.5. Update parameters. After determining the partial derivatives of errors for all model parameters according to formula (4), the Levenberg-Marquardt method [38] is used to update the parameters and minimise errors. The parameter update process is as follows:

$$\theta_{k+1} = \theta_k - \left(J^T J + \lambda I\right)^{-1} J^T e, \qquad (5)$$

where $\theta_k$ is the vector of current model parameters, $J$ is the Jacobian matrix calculated by formulas (3), $\lambda$ is the damping parameter, $I$ stands for the identity matrix, and $e$ represents the error vector.

The process formalised by formula (5) is repeated iteratively until convergence is achieved, i.e., until the changes in the parameters become insignificant or the errors reach a minimum value.

Errors for all points are given as the sum of the Euclidean distances between their calculated and measured coordinates. The total error is calculated as follows:

$$E = \sum_{i=1}^{N} \sqrt{\left(f\frac{X_i}{Z_i} + c_u - u_{i'}\right)^2 + \left(f\frac{Y_i}{Z_i} + c_v - v_{i'}\right)^2}, \qquad (6)$$

where $u_i = f\frac{X_i}{Z_i} + c_u$ and $v_i = f\frac{Y_i}{Z_i} + c_v$ are the calculated program coordinates along the $u$ and $v$ axes, respectively, $f$ is the focal length, $c_u$ and $c_v$ are the principal point coordinates.

Formula (6) reflects the differences between the calculated program coordinates and the actual measured values for each point, allowing for the assessment of model accuracy according to the established criteria for each UAV optical sensor separately.

## B. COMPOSITION OF VISUAL DATA STRUCTURES

Block 2 of the method performs the composition of visual data structures (see Fig. 2) according to the following steps.

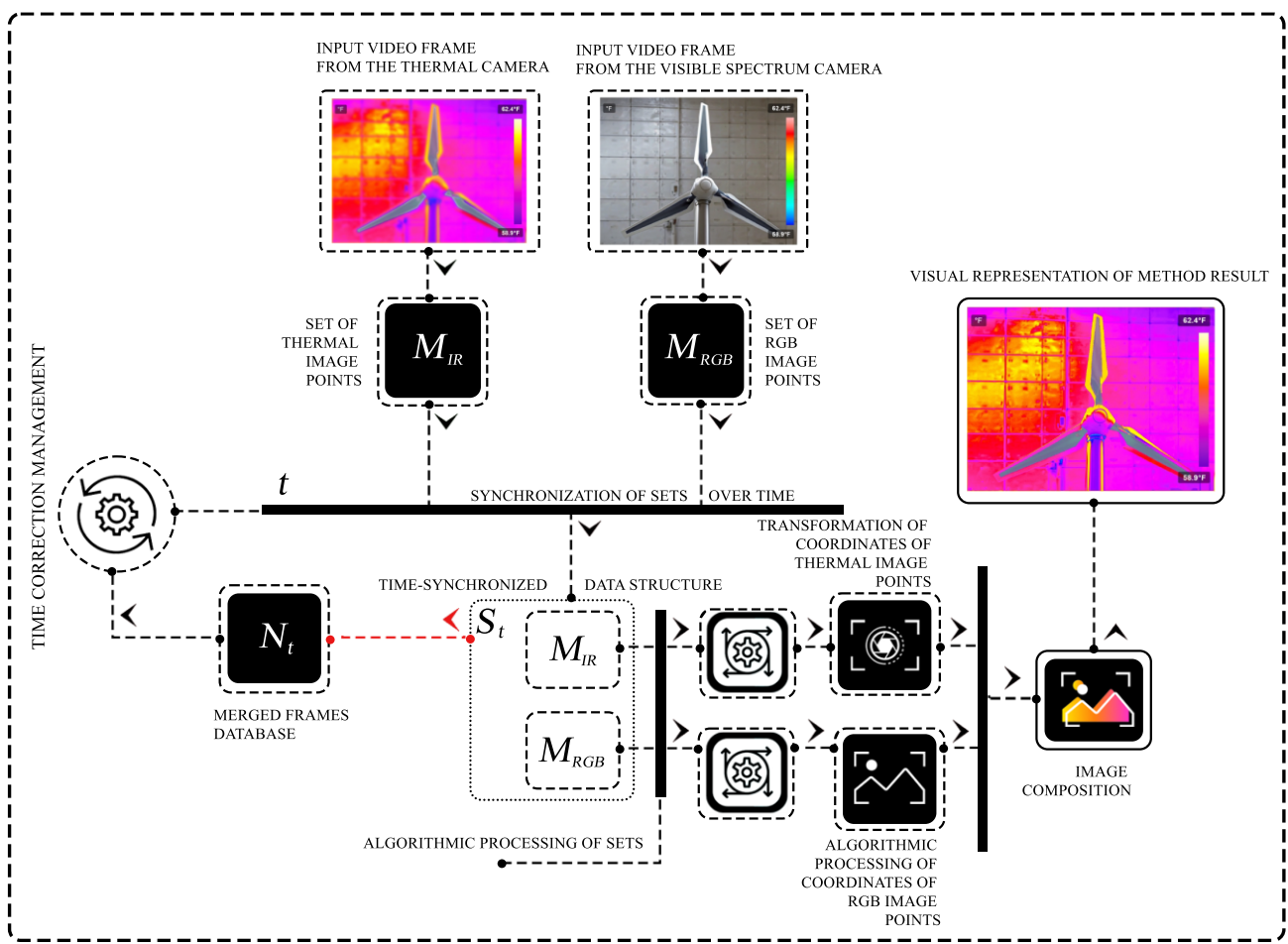

Figure 2. Scheme of the proposed visual data structure composition block for multispectral image composition: Input video frames from a UAV's thermal and visible spectrum cameras generate sets of thermal ($M_{IR}$) and RGB ($M_{RGB}$) image points. These sets are synchronised over time into a time-synchronised data structure ($S_t$), involving time correction, merging frames into a database ($N_t$), and algorithmic processing to transform image points' coordinates. This process results in a final merged thermal-RGB image, integrating different spectral data.

Step 2.1. Synchronise the set of allowable visual points. The input data for this step is the set of allowable visual points in the program coordinates formed in the previous block: $M_t = \left\{ B_{x,y,1}, B_{x,y,2}, \ldots, B_{x,y,n_{oy}} \right\}$, where $B_{x,y,n_{oy}}$ are the coordinates of these points on the image.

The primary task for the composition of the sets $M_t$ is their synchronisation in time. We denote the set of points from the thermal camera as $M_{t,IR} = \left\{ B_{IR,x,y,1}, B_{IR,x,y,2}, \ldots, B_{IR,x,y,n_{IR}} \right\}$, and the set of points from the RGB camera as $M_{t,RGB} = \left\{ B_{RGB,x,y,1}, B_{RGB,x,y,2}, \ldots, B_{RGB,x,y,n_{RGB}} \right\}$. The synchronised data structure is defined as follows:

$$S_t = \left\{ \left( ID_i, t, B_{IR,x,y,i}, B_{RGB,x,y,i} \right) \mid \left( ID_i, t, B_{IR,x,y,i}, B_{RGB,x,y,i} \right) \right\}, \tag{7}$$

where $ID_i$ is a unique identifier for each pair of points from the thermal camera and the RGB camera, $t$ is the synchronisation time, $B_{IR,x,y,i}$ are the coordinates of the point from the thermal camera, $B_{RGB,x,y,i}$ are the coordinates of the point from the RGB camera.

2.2. Determine key points. The next step in the composition of visual data structure by (7) is the identification of key points. For each pixel $p$ in the image, a circle of 16 pixels around it is selected. Let the intensity of the central pixel be $I_p$, and the intensity of an adjacent pixel be $I_n$, where $n = 1, 2, \ldots, 16$. A pixel $p$ is considered a key point if the intensity of at least 7 neighbouring pixels satisfies one of the conditions: $I_n > I_p + \tau$ or $I_n < I_p - \tau$, where $\tau$ is the threshold value. Formally, a pixel $p$ is a key point if:

$$\sum_{n=1}^{16} \left[ (I_n > I_p + \tau) \vee (I_n < I_p - \tau) \right] \geq 7, \tag{8}$$

where the symbol $\vee$ denotes logical "or".

2.3. Calculate the orientation of key points. To ensure rotation invariance in the proposed method, the moments method is used to calculate the orientation of key points calculated by the formula (8). For each such point $p$, the intensity moments in the area $p$ are calculated:

$$m_{10} = \sum_x \sum_y x \cdot I(x,y), \quad m_{01} = \sum_x \sum_y y \cdot I(x,y), \tag{9}$$

where $I(x,y)$ is the intensity of the pixel with coordinates $(x,y)$.

Based on these moments in formula (9), the orientation of the key point $\theta$ is calculated as follows:

$$\theta = \arctan\left( \frac{m_{01}}{m_{10}} \right). \tag{10}$$

The calculation of $\theta$, according to formula (10), allows identifying key points based on their orientation, ensuring rotation invariance of the image.

2.4. Create binary descriptors. The binary descriptor for each key point detected in the image represents the local structure of the image as a binary vector, which is resistant to changes in lighting, noise, and rotation. A set of pixel pairs $(p_a, p_b)$ is selected according to a defined scheme from the area around the key point. For each pair, the intensities $I(p_a)$ and $I(p_b)$ are compared. The descriptor $D$ for the key point $p$ is formed as a binary vector $D = \left\{ D(i) \right\}_{i=1}^n$, according to the following rule:

$$D(i) = \begin{cases} 1, & if \quad I\left(p_a^i\right) < I\left(p_b^i\right); \\ 0, & if \quad I\left(p_a^i\right) < I\left(p_b^i\right), \end{cases} \tag{11}$$

where $D$ is the binary descriptor vector, $i$ is the index of the pixel pair used for comparison, $\left(p_a^i\right)$ and $\left(p_b^i\right)$ are the coordinates of the $i$-th pair of pixels in the local area around the key point, and $I(p)$ is the intensity of the pixel at point $p$.

Step 2.5. Mark key points and matches. To visualise the results of the composition of two images, an approach is used that marks key points and lines connecting matching pairs. Let $M \subseteq \left\{ (i,j) \mid 1 \leq i \leq m, 1 \leq j \leq n \right\}$ be the set of indices of matching points. Then, for each pair $(i,j) \in M$, a key point $(x_i, y_i)$ is marked on image 1 and $(x_i', y_i')$ on image 2, along with a line between these points.

2.6. Transform key point coordinates. The transformation of key point coordinates is performed using the homography matrix $H$ [39], which is formalised as:

$$H = \begin{pmatrix} h_{11} & h_{12} & h_{13} \\ h_{21} & h_{22} & h_{23} \\ h_{31} & h_{32} & h_{33} \end{pmatrix}. \tag{12}$$

The homography matrix $H$, calculated by formula (12), determines all parameters of the projective transformation, namely translation, scaling, rotation, and perspective distortions. To compute the homography matrix $H$, corresponding pairs of points from the two images are used. Then, the projective transformation of the coordinates of the point $(x, y)$ from image 1 to the point $(x', y')$ from image 2 is carried out based on the homography matrix $H$ using the equations:

$$x' = \frac{h_{11}x + h_{12}y + h_{13}}{h_{31}x + h_{32}y + h_{33}}, \quad y' = \frac{h_{21}x + h_{22}y + h_{23}}{h_{31}x + h_{32}y + h_{33}}. \tag{13}$$

Let $(x_i, y_i)$ be the coordinates of the key points on image 1, and $(x_i', y_i')$ be the coordinates of the corresponding points on image 2, calculated by formula (13). Knowing such pairs of points $(x_i, y_i)$ and $(x_i', y_i')$, it is possible to transform the coordinates of points from one image into the coordinates of points on another image. The input data are two images $I_1$ and $I_2$, between which the matching key points are defined. A set of pairs of key points $(p_i, q_i)$ is used to compute the homography matrix $H$ for $i = 1, \ldots, N$, where $p_i = (x_i, y_i)$ are the coordinates of the key point on the first image, and $q_i = (x_i', y_i')$ are the coordinates of the corresponding key point on the second image.

At this step, the minimisation problem is solved, which consists of finding such a matrix $H$ that minimises the transformation error for all pairs of corresponding points. Let us define the error function, which is intended to calculate the distance between the corresponding points after applying the transformation:

$$E_i = |q_i - Hp_i|, \tag{14}$$

where $|\cdot|$ denotes the Euclidean norm.

The error function that generalises (14) for all pairs of points is defined as the sum of errors for each pair:

$$E(H) = \sum_{i=1}^{N} E_i = \sum_{i=1}^{N} |q_i - Hp_i|. \tag{15}$$

Thus, the problem of calculating the matrix $H$ reduces to minimising the error function from formula (15), namely

$$H = \arg\min_H \{E(H)\},$$

which ensures precise alignment of the two images.

2.7. Overlay the image. The transformation is carried out for the thermal camera image, as it has a lower resolution. Let $I_w$ be the transformed thermal image, and $I_2$ be the unchanged thermal image. To obtain the composition, a weighted overlay method [40] is used, which ensures a balance between detail and contrast:

$$R = \alpha \cdot I_w + \beta \cdot I_2 + \gamma, \tag{16}$$

where $\gamma$ is the offset constant for adjusting the overall brightness and contrast, $I_w(x, y)$ is the intensity of the transformed thermal image at the point $(x, y)$, and $I_2(x, y)$ is the intensity of the RGB image at the point $(x, y)$, coefficients $\alpha$ and $\beta$ are chosen to ensure a balance between detail and contrast; under the normalisation condition $- \alpha + \beta = 1$.

Equation (16) allows the overall intensity of the composite image $R$ to remain stable.

Next, the pixel intensity is adjusted during the weighted overlay of images.

Step 2.8. Calculate the average intensity. At this stage, the average intensity for each image is calculated. The average intensity $I_w$ for image 1 is determined as follows:

$$\overline{I_w} = \frac{1}{N} \sum_{(x,y) \in I_w} I_w(x, y), \tag{17}$$

where $N$ is the total number of pixels in the image $I_w$.

The average intensity $I_2$ for image 2 is determined as:

$$\overline{I_2} = \frac{1}{N} \sum_{(x,y) \in I_2} I_2(x, y). \tag{18}$$

Step 2.9. Determine the offset constant. The next step is to determine the offset constant $\gamma$ from formula (16), which is used to adjust the overall brightness and contrast of the composite image. The offset constant is calculated as the average of the average intensities (17) and (18) of both images using the formula:

$$\gamma = \frac{\overline{I_w} + \overline{I_2}}{2}.$$

Step 2.10. Optimise weight coefficients. The weight coefficients $\alpha$ and $\beta$ are determined through optimisation, which involves minimising the difference between the overlaid images to ensure the most accurate representation of both images in one composite. The optimisation problem is formulated as follows:

$$\begin{aligned}(\alpha, \beta) = \arg\min_{\alpha, \beta} \sum_{(x,y)} \big(\alpha \cdot I_w(x, y) + \\ + \beta \cdot I_2(x, y) - R(x, y)\big)^2.\end{aligned} \tag{19}$$

The process described by formula (19) finds such values of $\alpha$ and $\beta$ that minimise the sum of the squared differences between the intensities of the overlaid images and the final composite image $R$ from formula (17). This step ensures that the contribution of each image is balanced and best reflects their characteristics. The result of applying the proposed method to the composition of the RGB image and the thermal image is shown in Fig. 3.

Thus, the output of the composition method is a new image $R$, which combines data from two cameras, ensuring precise overlaying of information from both images. The image $R$ generated in this way allows for consideration of both thermal anomalies detected by the thermal camera and visual details obtained using the RGB camera.

## IV. RESULTS AND DISCUSSION

For training and testing the models, the Small-WTB-Thermal1 [36] dataset was used to train and test models, featuring 1000 thermal images of WTBs, split between 500 "healthy" and 500 "faulty" classes. Faulty images display defects like cracks, erosion, and holes. Each image was processed to a resolution of 320×320 pixels, with additional thermal-specific normalization and augmentations to enhance robustness for real-world conditions.

The practical significance of the proposed method was evaluated by solving the problem of detecting visual defects in WTB images and comparing the results of detection with state-of-the-art object detection models: Faster R-CNN [19], EfficientDet [20], YOLOv5 [21], and YOLOv8 [22]. All these models were trained on regular RGB images and fine-tuned on composite images created using the proposed method. The detection results are presented in Table 1.

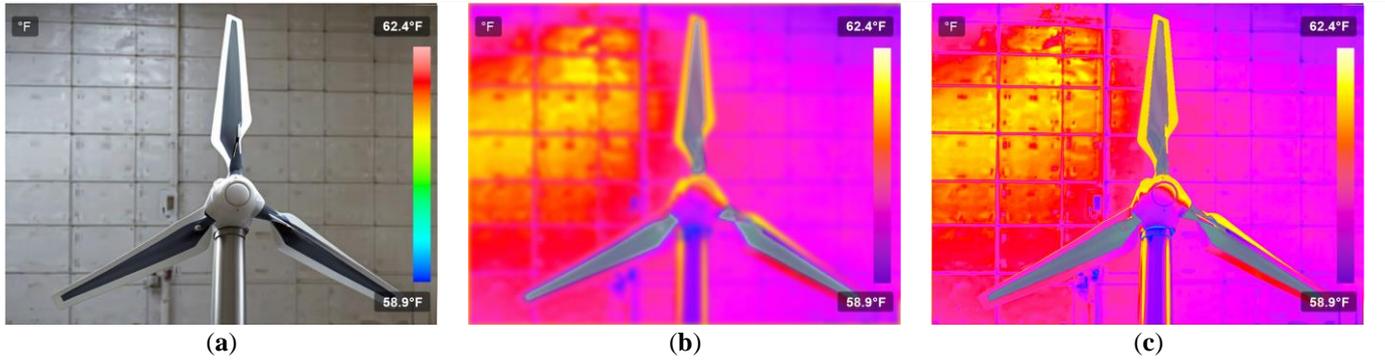

<div align="center">(a)            (b)            (c)</div>

Figure 3. A visual representation of the result of compositing two images: the original RGB image from the visible spectrum camera (a), the original thermal image from the thermal camera (b), and the resulting composite image $R$ (c) that integrates both spectral data sets for comprehensive visualisation.

**Table 1. The comparison of defect detection performance between RGB and composite images using four models.**

| Model | Image type | Number of false positives | Number of missed defects | Average image processing time, sec |
|---|---|---|---|---|
| Faster R-CNN [19] | RGB | 7 | 6 | 0.12 |
| | Composite | 4 | 3 | 0.14 |
| EfficientDet [20] | RGB | 9 | 8 | 0.08 |
| | Composite | 6 | 5 | 0.10 |
| YOLOv5 [21] | RGB | 8 | 7 | 0.05 |
| | Composite | 5 | 4 | 0.07 |
| YOLOv8 [22] | RGB | 6 | 5 | 0.04 |
| | Composite | 3 | 2 | 0.06 |

The results in Table 1 show that the proposed image composition method improves the performance of defect detection for all models. The number of false positives decreased from 8 to 5 for YOLOv5 and from 6 to 3 for YOLOv8. The number of missing defects also decreased: from 7 to 4 for YOLOv5 and from 5 to 2 for YOLOv8. Although the processing time slightly increased, the benefits in the reduction of false positives are significant.

Several performance statistical metrics, including accuracy, precision, recall, F₁-score, specificity, and sensitivity, were used to evaluate the effectiveness of the models on different images. The metrics' values are shown in Table 2.

The values in Table 2 show that the use of composite images significantly improves the performance of the models. For example, the accuracy of YOLOv5 increased from 0.88 to 0.92, and YOLOv8 increased from 0.91 to 0.95. Additionally, for the YOLOv8 model, precision increased from 0.89 to 0.94, recall from 0.85 to 0.92, and F₁-score from 0.87 to 0.93. Specificity increased from 0.90 to 0.95, and sensitivity from 0.84 to 0.91.

According to Fig. 4, the use of composite images allows for detecting more small and unnoticeable defects than regular images. In the RGB images (Fig. 4(a) and (c)), some defects remain unnoticed or difficult to recognise, while in the thermal images (Fig. 4(b) and (d)), the same defects are clearly highlighted due to differences in the thermal characteristics of the materials. Thus, composite images that combine information from both types of images significantly improve the ability of models to detect defects, contributing to more accurate diagnostics of the condition of wind turbines.

Despite the advancements above, the proposed method has several limitations, including increased processing time and sensitivity to environmental conditions, and it could be a subject for future research.

## V. CONCLUSIONS

In this study, we presented a novel method for enhancing defect detection on WTBs by integrating thermal and RGB imagery captured by unmanned aerial vehicles. Our multispectral image composition method effectively combines thermal and RGB images, leading to significant improvements in defect detection performance: the YOLOv8 model's accuracy increased from 91% to 95%, precision from 89% to 94%, recall from 85% to 92%, and F1-score from 87% to 93%. Additionally, the number of false positives decreased from 6 to 3, and missed defects reduced from 5 to 2. Despite these promising results, the proposed approach has limitations, including increased computational complexity due to additional image processing steps and potential challenges in aligning thermal and RGB images under varying environmental conditions.

Future work will focus on optimizing the method to reduce processing time, enhancing image alignment techniques, and integrating the approach into real-time UAV inspection workflows to further improve the efficiency and applicability of WTB maintenance.

**Table 2. The comparison of defect detection performance between RGB and composite images by accuracy, precision, recall, F₁-score, specificity, and sensitivity.**

| Model | Image type | Accuracy | Precision | Recall | F₁-score | Specificity | Sensitivity |
|---|---|---|---|---|---|---|---|
| Faster R-CNN [19] | RGB | 0.89 | 0.87 | 0.84 | 0.85 | 0.91 | 0.83 |
| | Composite | 0.93 | 0.892 | 0.90 | 0.91 | 0.94 | 0.89 |
| EfficientDet [20] | RGB | 0.87 | 0.85 | 0.82 | 0.83 | 0.89 | 0.81 |
| | Composite | 0.91 | 0.90 | 0.87 | 0.88 | 0.93 | 0.86 |
| YOLOv5 [21] | RGB | 0.88 | 0.86 | 0.83 | 0.84 | 0.90 | 0.82 |
| | Composite | 0.92 | 0.91 | 0.89 | 0.90 | 0.94 | 0.88 |
| YOLOv8 [22] | RGB | 0.91 | 0.89 | 0.85 | 0.87 | 0.90 | 0.84 |
| | Composite | 0.95 | 0.94 | 0.92 | 0.93 | 0.95 | 0.91 |

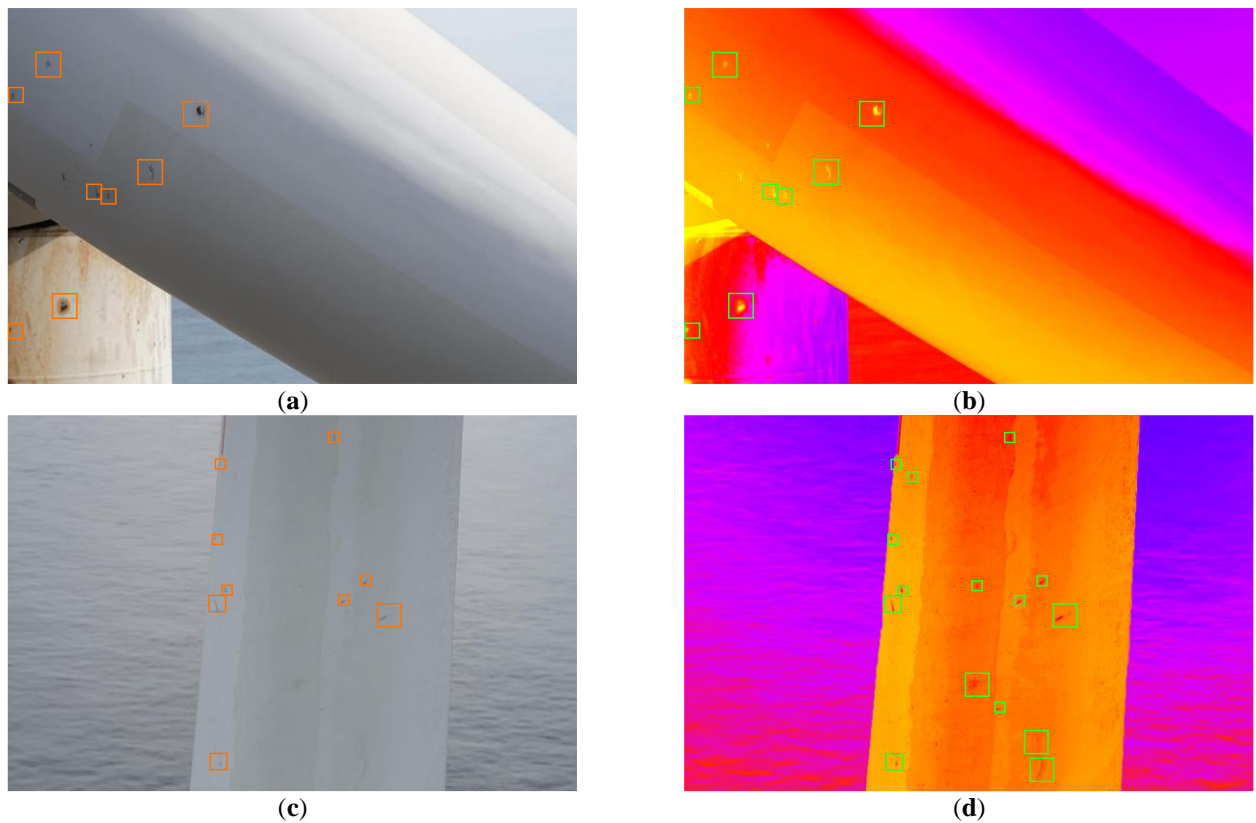

**Figure 4.** The visual results of defect detection on WTBs using both RGB and composite images. Panels (a) and (c) show the original RGB images with defects highlighted in orange, while panels (b) and (d) present the composite images with defects highlighted in green, illustrating the enhanced visibility and detection accuracy achieved by integrating thermal and RGB data.

···